%% file: main.tex
\documentclass[letterpaper]{article} 
\usepackage{aaai2027}  
\usepackage[hyphens]{url}  
\usepackage{graphicx} 
\usepackage{natbib}  
\usepackage{caption} 
\usepackage{algorithm}
\usepackage{algorithmic}

\usepackage{newfloat}
\usepackage{listings}
\DeclareCaptionStyle{ruled}{labelfont=normalfont,labelsep=colon,strut=off} 
\floatstyle{ruled}
\newfloat{listing}{tb}{lst}{}
\floatname{listing}{Listing}

\usepackage{booktabs}

\usepackage{amsmath}
\usepackage{amssymb}
\usepackage{amsfonts}
\usepackage{arydshln}
\usepackage{bm}
\usepackage{multirow}
\usepackage{subcaption}

\newcommand{\dataset}{TeleRCA}

\newcommand{\Dhist}{\mathcal{D}_{\mathrm{hist}}}
\newcommand{\RCA}{\textsc{RCAEngine}}
\newcommand{\CD}{\textsc{CausalDiscovery}}
\newcommand{\TopoOrder}{\textsc{TopologicalOrder}}
\newcommand{\ApplyG}{\textsc{ApplyGraphUpdates}}
\newcommand{\indic}{\mathbf{1}}
\newcommand{\Dalign}{\mathcal{D}_{\mathrm{align}}}
\newtheorem{assumption}{Assumption}

\title{EvoCause: LLM-Guided Evolution of Causal Graphs for Root Cause Analysis}
\author{
    Lei Zan\textsuperscript{\rm 1},
    Keli Zhang\textsuperscript{\rm 1},
    Shifeng Xie\textsuperscript{\rm 1},
    Jiale Zheng\textsuperscript{\rm 1},
    Zehao Xiao\textsuperscript{\rm 1},\\
    Zhiwei Dong\textsuperscript{\rm 1},
    Ke Zhang\textsuperscript{\rm 1},
    Ruichu Cai\textsuperscript{\rm 2},
    Malik Tiomoko\textsuperscript{\rm 1},
    Lujia Pan\textsuperscript{\rm 1}
}

\affiliations{ \textsuperscript{\rm 1}Huawei Noah's Ark Lab\\ \textsuperscript{\rm 2}Guangdong University of Technology, Peng Cheng Laboratory\\ \{xdzanlei, zxiao4ai, cairuichu\}@gmail.com,\\ \{zhangkeli1, zhengjiale2, zhangke200, malik.tiomoko, panlujia\}@huawei.com,\\ shifeng.xie@telecom-paris.fr, zhiwei.dong@etu.sorbonne-universite.fr }

\begin{document}

\maketitle

\begin{abstract}
Modern telecommunication, cloud, and microservice systems emit correlated alarm cascades when components fail. Root cause analysis (RCA) aims to identify the small set of alarms that initiate each cascade. A common approach learns a causal graph from observational logs and predicts all zero-in-degree alarms in each incident-induced subgraph. However, the learned graph remains fixed and cannot benefit from expert diagnoses of historical incidents. We close this loop with \textbf{EvoCause}. Expert labels constrain which alarms should be source nodes but do not specify the edge edits needed to satisfy those constraints. EvoCause uses a large language model (LLM) to propose semantically plausible graph edits, while deterministic code validates node identities and acyclicity and retains the best graph on a labeled alignment set. At test time, the refined graph alone produces transparent predictions without an LLM call. We also release \dataset, an expert-annotated benchmark from a production telecommunication network containing $485{,}681$ alarm events spanning $194$ alarm types over $5{,}621$ resources. On synthetic data, EvoCause initialized with the PC causal discovery algorithm outperforms the unrefined PC baseline, raising Node F1, Case EM, and Graph F1 by $11.59$, $9.40$, and $4.59$ percentage points, respectively, while reducing nSHD by $0.2379$. On TeleRCA, replacing human-readable alarm titles with anonymous identifiers lowers Node F1 and Case EM by $6.12$ and $8.04$ percentage points, respectively, indicating that alarm-name information contributes to graph refinement.
\end{abstract}


\input{Sub_sections/Introduction}
\input{Sub_sections/Sota}
\input{Sub_sections/Methodology}
\input{Sub_sections/Experiments}
\input{Sub_sections/Conclusion}

\bibliography{references}


\clearpage
\appendix
\input{Sub_sections/Appendix}

\end{document}

%% file: Sub_sections/Introduction.tex
\section{Introduction}
Modern telecommunication, cloud, and microservice systems comprise many tightly coupled components, so a single fault can trigger a cascade of correlated alarms, known as an ``alert storm''~\citep{zhao2020understanding}. In this context, root cause analysis (RCA) is the task of identifying, among these alarms, the smallest set of actionable events whose remediation actually resolves the incident~\citep{pmlr-v206-assaad23a}. Rapid diagnosis is critical because major outages can disrupt many dependent services~\citep{bajak2021aws}.

A common causal approach represents alarm types and their triggering relations as a directed acyclic graph (DAG) and predicts all zero-in-degree alarms in the subgraph induced by each incident. This framing provides traceable predictions and generalizes across incidents by encoding triggering relations rather than memorized cases. 

Constructing such a graph manually is impractical at scale, so causal discovery algorithms learn it from observational data~\citep{assaad2022survey,spirtes2000}. However, their assumptions and dependence on clean data can produce spurious or missing edges in production alarm streams~\citep{aitbachir2023casestudiescausaldiscovery}.

Crucially, existing methods ignore a complementary source of information that could correct these structural errors, namely the historical incident logs curated by Site Reliability Engineering (SRE) teams, which pair observed alarm sequences with expert-labeled root causes. These diagnostic labels are sparse and expensive to obtain. More importantly, they provide task-level constraints rather than direct edge supervision: they can rule out graphs whose incident-induced source nodes disagree with expert diagnoses, but they generally do not identify a unique ground-truth DAG.

This feedback is indirect and underdetermined because expert labels specify which alarms should be sources but not which edge additions, removals, or reversals should enforce them without degrading predictions for overlapping incidents. Language models are useful because they can jointly interpret constraints from multiple incidents, the current graph, optimization history, and, when available, information from alarm titles. We therefore propose \textbf{EvoCause} (Evolve Causal Graph), which uses an LLM to rank a small set of semantically plausible edits from this combinatorial space rather than to predict the root cause directly. Deterministic procedures validate every edit, reject cyclic candidates, and retain the graph with the best alignment-set performance. The graph is the only optimized state, whereas its derived topological order is used only as prompt context rather than as a prediction rule. We evaluate the contribution of alarm-title information on TeleRCA by replacing only the titles supplied to the LLM with fixed anonymous identifiers while preserving the graph structure and incident data. Our contributions are summarized as follows:

\begin{itemize}

  
 \item On the methodological side, we formulate incident-level expert feedback as \emph{source-node constraints} on incident-induced subgraphs and characterize the corresponding set of root-cause-consistent causal graphs. Based on this formulation, we propose \textbf{EvoCause}, a discovery-agnostic causal graph refinement framework that integrates LLM-guided semantic graph editing with deterministic structural validation and supervision-driven graph selection. EvoCause can refine graphs produced by different causal discovery algorithms without modifying their internal learning procedures, while ensuring that the resulting graphs remain structurally valid and consistent with expert RCA annotations.

\item On the data side, we release \textbf{TeleRCA}, a large-scale, expert-labeled RCA benchmark derived from a real-world production telecommunication network. The dataset comprises $10{,}922$ incidents, $485{,}681$ alarm events, and $194$ alarm types across thousands of network resources. Its combination of production alarm sequences and incident-level expert root-cause annotations provides a complementary testbed for advancing and evaluating event-sequence-based RCA methods.

  \item On the experimental side, we evaluate both RCA (Node F1 and Case EM) and graph reconstruction (Graph F1 and nSHD) on ten synthetic DAGs, together with RCA performance on TeleRCA. EvoCause improves all three discovery backbones. A controlled name-anonymization ablation on TeleRCA evaluates the contribution of human-readable alarm-title information, while additional backbone comparisons on synthetic data assess sensitivity to the choice of LLM.


\end{itemize}

%% file: Sub_sections/Sota.tex
\section{Related Work}

\paragraph{Causal graphs and expert feedback.} Observational data generally identify a causal DAG only up to a Markov equivalence class under some standard assumptions. A completed partially directed acyclic graph (CPDAG) represents this class. Directed edges are shared by all compatible DAGs, while undirected edges remain ambiguous~\citep{chickering2002learning}. Consistent edge-orientation background knowledge can orient some ambiguous edges and narrow the class~\citep{fang2025background}. Propagating all implied orientations yields a maximally oriented partially directed acyclic graph (MPDAG), which represents the DAGs consistent with both the data and the background knowledge~\citep{perkovic2020identifying,guo2022mpdag}. Expert root-cause labels instead constrain source sets in incident-induced subgraphs and may require skeleton changes, so EvoCause searches a root-cause-consistent graph set that can span multiple equivalence classes. Interventional equivalence would apply only if incidents were known interventions~\citep{hauser2012imec}. In telecommunication RCA, \citet{zhang2021influence} combine Hawkes-process discovery, conditional-independence tests, propagation embeddings, and influence maximization. CCCM~\citep{zhang2024cluster} introduces cluster-aware discovery for base-station alarms, while TTCTH~\citep{li2025ttcth} combines topological--temporal representation learning with a Hawkes-process objective. For microservices, CIRCA~\citep{ikram2022neurips} treats failures as interventions, RUN~\citep{lin2024run} combines neural Granger causality with contrastive learning, and CausIL~\citep{causil2023} incorporates domain knowledge into instance-level causal graphs. 
For threshold-based IT systems, \citet{zan2024onthefly} combine offline causal discovery with online subgraph traversal. 
APGNN~\citep{jiang2023apgnn}, Chain-of-Event~\citep{yao2024coe}, and Groot~\citep{wang2021groot} construct event or alarm graphs for diagnosis using learned associations, interpretable edge parameters, or operator rules. These methods provide graph-learning and localization mechanisms but generally do not repeatedly revise an alarm-type DAG using accumulated expert root sets. HRLHF~\citep{wang2023hrlhf} does query engineers during dependency-graph discovery, whereas EvoCause uses previously labeled incidents as an offline task signal for refining a graph produced by any discovery backbone.
\paragraph{LLM-based RCA and graph refinement.} LLM-based RCA systems have been studied for recommending root causes and mitigations from cloud-incident text~\citep{ahmed2023icse}. RCACopilot~\citep{chen2024eurosys} combines incident-specific diagnostic collection with classification and explanation. RCAgent~\citep{wang2024rcagent} and ReAct-based systems~\citep{roy2024fse} use tools and retrieval, while OpenRCA~\citep{xu2025openrca} and Flow-of-Action~\citep{pei2025wow} provide agent-oriented evaluation or procedure-constrained diagnosis. These approaches mainly predict, explain, or plan actions for an individual incident, so their outputs need not update a shared causal model. Cloud Atlas~\citep{xie2024cloudatlas} is closer to reusable graph construction because it synthesizes causal graphs from documentation, telemetry, and deployment feedback and then validates them with data. Outside RCA, CAMA refines a mathematical causal graph using question-answer feedback~\citep{zan2026cama}. EvoCause instead refines an externally discovered alarm graph using expert root-cause labels and performs test-time inference with the refined graph alone. The LLM proposes semantic candidates, while deterministic procedures validate node identities and acyclicity, select the best graph, and perform all test-time inference.
\paragraph{Datasets.} The telecom data used by \citet{zhang2021influence} provide root-cause labels for 6,000 sampled alarm transactions, with causal-edge annotations covering 15 of 78 alarm types. THPs~\citep{cai2024thps} evaluates event-sequence structure learning rather than incident-level root localization. OpenRCA contains 335 software failures and more than 68~GB of logs, metrics, and traces~\citep{xu2025openrca}. In comparison, \dataset{} targets alarm-sequence RCA and provides expert root-cause annotations for 10,922 production incidents. It does not provide a ground-truth propagation graph, so we use it to evaluate RCA rather than exact graph reconstruction.

%% file: Sub_sections/Methodology.tex
\section{EvoCause}
 
This section formulates RCA on alarm cascades and presents EvoCause, which has learning and inference stages. Section~\ref{sec:setup} defines the problem and assumptions, Section~\ref{sec:identifiability} analyzes identifiability under root-cause feedback, Section~\ref{sec:overview} summarizes the framework, Section~\ref{sec:learning} describes graph construction and label-feedback alignment, and Section~\ref{sec:inference} explains root cause inference.
 
\subsection{Problem Setup}
\label{sec:setup}
 
\paragraph{Dataset Description.}
We consider a telecommunication network that generates alarms from a finite set of alarm types, denoted by $\mathbf{A}$. The dataset contains $m$ incident sequences, $\mathcal{S}=\{\mathbf{s}_i\}_{i=1}^{m}$, collected over a global observation window $\mathbf{T}$. Each incident sequence $\mathbf{s}_i$ corresponds to a single incident and is represented as
\begin{equation} 
\mathbf{s}_i=\left\{(a_j,t_j,l_j)\right\}_{j=1}^{n_i}. 
\end{equation}
where $n_i$ denotes the number of alarm events in incident $\mathbf{s}_i$. Each triplet represents one alarm event: $a_{j} \in \mathbf{A}$ represents the alarm type, $t_{j} \in \mathbf{T}_i$ represents its timestamp, and $l_{j} \in \{0,1\}$ is an expert-provided annotation indicating whether the event is a root cause of the corresponding incident. Specifically, $l_{j}=1$ indicates a root-cause event, whereas $l_{j}=0$ indicates a non-root-cause event. The corresponding incident-level expert root set is $y_i=\{a_j:(a_j,t_j,l_j)\in\mathbf{s}_i,\ l_j=1\}$, where repeated occurrences of the same alarm type are included only once. We further assume that the triggering relations among alarm types in $\mathbf{A}$ remain invariant throughout the observation window.

\paragraph{Alarm propagation graph.} We represent the triggering relations among the alarm types $\mathbf{A}$ with a directed acyclic graph $\mathcal{G} = (\mathbf{A}, \mathbf{E})$, which we call an \emph{alarm propagation graph} (APG). Its nodes are alarm types, and an edge $(a_i,a_j)\in\mathbf{E}$ means that an occurrence of $a_i$ may trigger $a_j$. For an incident, RCA restricts $\mathcal{G}$ to the observed alarm types and predicts every zero-in-degree node in the induced subgraph as a root cause. The graph is the only persistent structural state. When a linear representation is useful for prompting, we deterministically derive a topological order $O(\mathcal{G})=\TopoOrder(\mathcal{G})$, in which $a_i$ precedes $a_j$ for every edge $(a_i,a_j)\in\mathbf{E}$. This order provides a compact linear view of the graph for the LLM. It is derived from $\mathcal{G}$ and is neither optimized nor used for root-cause prediction.

\paragraph{Incident-level source operator.}
Let $V_i\subseteq\mathbf{A}$ denote the distinct alarm types observed in incident $\mathbf{s}_i$. For a candidate APG $\mathcal{G}$, define
\begin{equation}
  \operatorname{Src}_{\mathcal{G}}(V_i)
  = \left\{v\in V_i:
  \operatorname{Pa}_{\mathcal{G}}(v)\cap V_i=\varnothing\right\}.
  \label{eq:source-operator}
\end{equation}

EvoCause predicts the complete set $\operatorname{Src}_{\mathcal{G}}(V_i)$ and neither ranks these nodes nor uses the ground-truth number of roots.

\paragraph{Learning Objective.}
A conventional causal approach obtains such a graph from the statistical
regularities of the observed alarm events alone, without using the expert
labels. A causal discovery algorithm $\CD(\cdot)$ maps the alarm types and
their timestamps to a DAG,
\begin{equation}
  \hat{\mathcal{G}} = \CD\big(\{(a_j, t_j)\}\big),
\end{equation}
and root causes are then localized by a procedure $h(\cdot)$ that assigns a
binary label to every event,
\begin{equation}
  \big[\hat{l}_1, \dots, \hat{l}_{n_i}\big]
  \;=\;
  h\!\left(\hat{\mathcal{G}},\, \{(a_j, t_j)\}_{j=1}^{n_i}\right).
\end{equation}
In this pipeline, the expert labels $l_j$ never enter the construction of $\hat{\mathcal{G}}$, because the graph is recovered from event statistics alone.
 
The core objective is to use expert-annotated incidents to refine the graph so that it recovers complete root sets more accurately. We split the incidents into an offline training set $\mathcal{S}_{\mathrm{off}}$ and a held-out test set $\mathcal{S}_{\mathrm{on}}$. For causal discovery, let $\pi_{\mathrm{obs}}(\mathbf{s}_i)=\{(a_j,t_j)\}_{j=1}^{n_i}$ denote the label-free projection of incident $\mathbf{s}_i$, and define $\Dhist=\{\pi_{\mathrm{obs}}(\mathbf{s}_i):\mathbf{s}_i\in\mathcal{S}_{\mathrm{off}}\}$. The initial graph is learned from $\Dhist$, while the labeled alignment set $\Dalign$ is used for graph refinement and model selection. All candidate-graph decisions use $\Dalign$, and $\mathcal{S}_{\mathrm{on}}$ is evaluated only once after selection. We select
\begin{equation} \mathcal{G}^{*}=\arg\max_{\mathcal{G}}\mathbb{E}_{(\mathbf{s}_i,y_i)\sim\Dalign}\left[\indic\{\RCA(\mathbf{s}_i,\mathcal{G})=y_i\}\right], \label{eq:objective} \end{equation}
where $y_i$ is the complete expert root set and $\indic\{\cdot\}$ is the indicator function. 

\paragraph{Root-cause-consistent graph set.}
Let $\mathcal{D}_{L}=\{(V_i,y_i)\}_{i=1}^{m}$ denote a set of labeled incidents, where $y_i\subseteq V_i$ is the expert-provided root set for incident $i$. We define the \emph{root-cause-consistent} (RCC) graph set as
\begin{equation}
  \mathcal{C}_{\mathrm{RCC}}(\mathcal{D}_{L})
  = \left\{\mathcal{G}\in\operatorname{DAG}(\mathbf{A}):
  \operatorname{Src}_{\mathcal{G}}(V_i)=y_i,\ \forall i\right\}.
  \label{eq:rcc-set}
\end{equation}
More generally, two graphs are RCA-equivalent on $\mathcal{D}_{L}$ if they produce the same source set for every labeled incident. Unlike a Markov equivalence class, the RCC set may contain graphs with different skeletons and conditional-independence structures. When exact consistency cannot be achieved because of label noise, a restricted search space, or optimization limitations, we quantify inconsistency using the empirical disagreement
\begin{equation}
  \mathcal{L}_{\mathrm{RCA}}(\mathcal{G},\mathcal{D}_{L})
  = \frac{1}{m}\sum_{i=1}^{m}\indic\!\left\{\operatorname{Src}_{\mathcal{G}}(V_i)\neq y_i\right\}.
  \label{eq:rcc-loss}
\end{equation}
The objective in~\eqref{eq:objective} then selects the visited graph with the smallest empirical disagreement, using Node F1 to break ties.

\paragraph{Assumptions.} Let $\mathcal{K}$ denote the resource--time contexts and let $\mathcal{G}^{(c)}$ be the alarm-type causal graph in context $c$.
\begin{assumption}[Causal consistency]
A shared DAG $\mathcal{G}^{\dagger}$ satisfies $\mathcal{G}^{(c)}=\mathcal{G}^{\dagger}$ for every $c\in\mathcal{K}$, corresponding to causal stationarity across contexts~\citep{assaad2022survey}.
\end{assumption}
\begin{assumption}[Causal sufficiency]
No latent alarm event is a common cause of two observed alarm types~\citep{spirtes2000,assaad2022survey}.
\end{assumption}
\begin{assumption}[Expert-label consistency]
Every alignment label satisfies $y_i=\operatorname{Src}_{\mathcal{G}^{\dagger}}(V_i)$.
\end{assumption}
These assumptions make the feedback well defined but do not make the RCC set unique. Violations of Assumptions 1 or 2 make the APG a task-effective summary rather than the unique data-generating graph. Let $\widetilde{\mathcal{D}}_L=\{(V_i,\widetilde{y}_i)\}_{i=1}^{m}$ denote a corrupted version of $\mathcal{D}_L=\{(V_i,y_i)\}_{i=1}^{m}$, where $\widetilde{y}_i\neq y_i$ for at most $q$ incidents. Then, for any fixed graph,
\begin{equation}
\left|\mathcal{L}_{\mathrm{RCA}}(\mathcal{G},\widetilde{\mathcal{D}}_L)-\mathcal{L}_{\mathrm{RCA}}(\mathcal{G},\mathcal{D}_L)\right|\leq\frac{q}{m}.
\label{eq:label-noise-bound}
\end{equation}

Each proposal considers a batch of incidents, and candidate graphs are scored on the full alignment set. A few isolated errors therefore have bounded direct influence, although systematic errors can still alter the proposal path and selected graph. EvoCause consequently targets an RCA-effective graph with low empirical disagreement rather than unique DAG recovery.

\subsection{Identifiability under Root-Cause Feedback}
\label{sec:identifiability}

\paragraph{Why root labels do not generally identify one DAG.}
For every labeled root $r\in y_i$, consistency requires $\operatorname{Pa}_{\mathcal{G}}(r)\cap V_i=\varnothing$. For every labeled non-root $v\in V_i\setminus y_i$, however, it only requires $\operatorname{Pa}_{\mathcal{G}}(v)\cap V_i\neq\varnothing$. The latter is a disjunctive constraint: at least one observed parent must exist, but the label does not identify which parent is responsible. Root-cause feedback therefore generally leaves multiple admissible DAGs.

\paragraph{Proposition 1 (Monotone refinement).}
Let $\mathcal{C}_m$ be the RCC set induced by the first $m$ labeled incidents. Then
\begin{equation}
  \mathcal{C}_{m+1}\subseteq\mathcal{C}_m.
\end{equation}
Indeed, the $(m+1)$-st label adds one source-set constraint, giving $\mathcal{C}_{m+1}=\mathcal{C}_m\cap\{\mathcal{G}:\operatorname{Src}_{\mathcal{G}}(V_{m+1})=y_{m+1}\}$. Thus additional consistent feedback can only eliminate candidate graphs.

\paragraph{Proposition 2 (Training-set RCA invariance).}
All graphs in $\mathcal{C}_{\mathrm{RCC}}(\mathcal{D}_{L})$ return the same expert root set on every incident in $\mathcal{D}_{L}$, although they may disagree on unseen incidents and may contain different edge sets. Consequently, training-label consistency alone does not determine which admissible graph will generalize best.

\paragraph{A sufficient condition for uniqueness.}
Suppose that the true APG is a DAG, its undirected skeleton is known, expert labels are noiseless, and \emph{root-separating coverage} holds: for every adjacency $\{u,v\}$ in the skeleton, the labeled set contains an edge-isolating incident whose relevant alarm-type set is $V_i=\{u,v\}$. For such an incident, the root set is $\{u\}$ if and only if the edge is $u\rightarrow v$, and it is $\{v\}$ if and only if the edge is $v\rightarrow u$. Every skeleton edge is therefore uniquely oriented, so the RCC set restricted to the known skeleton is a singleton. This condition is sufficient but intentionally strong and need not hold in \dataset. Without it, EvoCause targets task-level identification by selecting an RCA-effective representative rather than claiming recovery of the unique data-generating DAG.
 
\subsection{Overview of EvoCause}
\label{sec:overview}
 
\begin{figure*}[t]
  \centering
  \includegraphics[width=0.66\linewidth]{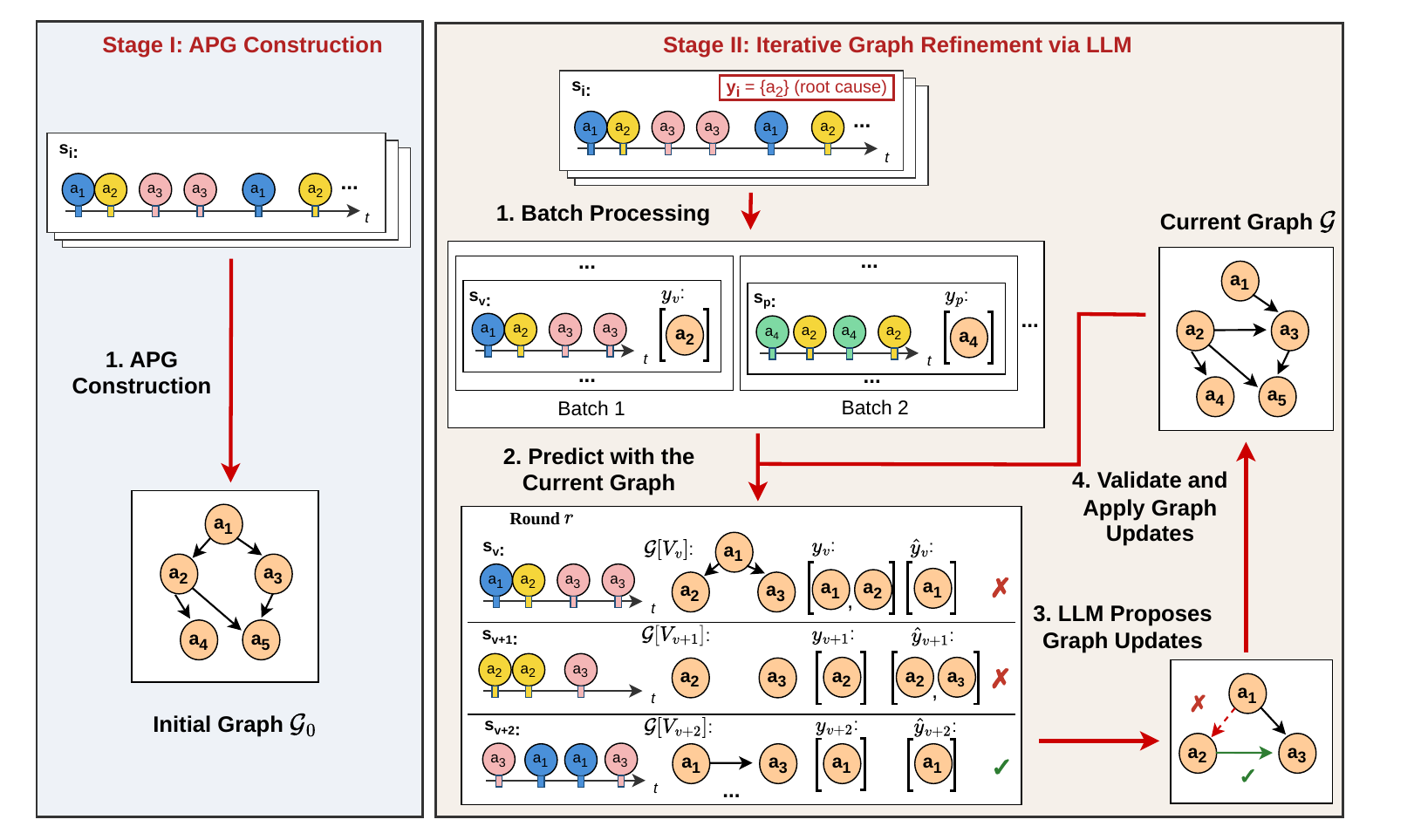}
  \caption{Overview of EvoCause. Stage~I learns an initial alarm propagation graph $\mathcal{G}_0$ from historical alarm sequences. In Stage~II, the current graph predicts the complete zero-in-degree source set for each expert-labeled incident in a batch. Prediction--label mismatches guide the LLM to propose edge additions, removals, or reversals. Deterministic code validates node identities and acyclicity, applies valid edits, and retains the graph with the best performance on the alignment set. Test-time RCA uses only the selected graph.}
  \label{fig:overview}
\end{figure*}

EvoCause comprises a \emph{learning stage} and an \emph{inference stage} (Figure~\ref{fig:overview}). In the learning stage, EvoCause first builds an initial alarm propagation graph $\mathcal{G}_0$ from historical alarm sequences using off-the-shelf causal discovery, and then refines it by aligning it with the downstream RCA task through LLM-based label feedback. In the inference stage, a new incident is resolved by running a transparent graph-based procedure on the refined graph $\mathcal{G}^{*}$.

\subsection{Learning Stage}
\label{sec:learning}
 
The learning stage has two components, namely (1) constructing an initial alarm propagation graph from historical sequences, and (2) aligning it with the RCA task through LLM-based label feedback.

\paragraph{Stage~I: Initial APG construction.}
The first stage constructs an initial graph from the label-free historical alarm sequences $\Dhist$:
\begin{equation} 
\mathcal{G}_0=\CD(\Dhist). 
\end{equation}
Here $\CD(\cdot)$ is any algorithm that returns a DAG over alarm types. We instantiate it with PC~\citep{spirtes2000}, NOTEARS~\citep{zheng2018notears}, and THPs~\citep{cai2024thps}, which have different inductive biases. EvoCause treats discovery as a black box. At each refinement round, a deterministic topological order $O(\mathcal{G})=\TopoOrder(\mathcal{G})$ is derived from the current DAG solely to give the LLM a compact linearization of causal flow. It is not a second model state, and root-cause prediction depends only on zero in-degree in the incident-induced subgraph.
 
\paragraph{Stage~II: APG alignment via label feedback.} 
The second stage refines $\mathcal{G}_0$ into $\mathcal{G}^{*}$ using labeled alignment cases $\Dalign=\{(\mathbf{s}_i,y_i)\}_{i=1}^{N}$, where $\mathbf{s}_i$ is an incident sequence and $y_i\subseteq V_i$ is its complete expert root set. Following the objective in~\eqref{eq:objective}, EvoCause searches for a graph that maximizes Case EM on the alignment set.

To optimize~\eqref{eq:objective}, we adopt a batch-based iterative optimization procedure over $N_e$ epochs. At the start of epoch $e$, $\Dalign$ is randomly permuted to obtain $\Dalign^{(e)}$ and then partitioned into batches of size $N_b$. Each batch triggers one refinement round with four steps.

\emph{Step 1: Batch processing.} A batch $\mathbf{B}_k$ of $N_b$ labeled cases is drawn from the current epoch's partition. Batching amortizes the cost of LLM invocation and allows each refinement round to aggregate feedback across multiple incidents, reducing sensitivity to any single case.

\emph{Step 2: Predict with the current graph.} For every case $(\mathbf{s}_i, y_i) \in \mathbf{B}_k$ we run a graph-based RCA procedure on the current graph,
\begin{equation}
  \hat{y}_i = \RCA(\mathbf{s}_i, \mathcal{G}),
\end{equation}
which restricts $\mathcal{G}$ to the alarm types observed in $\mathbf{s}_i$ and returns all zero-in-degree nodes of the incident-induced subgraph. The complete set of such source alarm types is predicted, and the ground-truth number of roots is never provided to the procedure. This yields a per-case triple $(\mathbf{s}_i,\hat{y}_i,y_i)$ whose mismatches are the discrepancies that the next step must explain.

\emph{Step 3: LLM proposes graph edits.} We aggregate the per-case triples into a batch-level evidence set
\begin{equation}
  \mathbf{E} = \{(\mathbf{s}_i, \hat{y}_i, y_i)\}_{i\in\mathbf{B}_k},
\end{equation}
and present $\mathbf{E}$, together with the current graph $\mathcal{G}$, its derived order $O(\mathcal{G})$, and an optimization history $\mathbf{H}$ (defined below), to a large language model $\mathcal{M}$ through an update prompt $p_u$. The LLM proposes only targeted graph edits
\begin{equation}
  \Delta\mathcal{G}
  \;=\;
  \mathcal{M}\!\left(\mathcal{G}, O(\mathcal{G}), \mathbf{E}, \mathbf{H}, p_u\right),
\end{equation}
where $\Delta\mathcal{G}$ is a set of directed-edge edits using the operations $\{\emph{add},\emph{remove},\emph{reverse}\}$. The LLM response contains only graph updates. The derived topological order $O(\mathcal{G})$ can neither be edited nor returned.

The LLM is a proposal model, not a causal oracle or a constraint solver. A labeled source-set error creates a disjunctive constraint: several edits may repair the current incident, yet those edits can have different effects on other incidents that share alarm types. Exhaustively testing all directed additions, removals, and reversals is quadratic in the number of alarm types per round before multi-edit combinations are considered, while a purely random proposer ignores both alarm semantics and cross-incident evidence. EvoCause uses the LLM to compress this search into a small set of proposals by jointly interpreting human-readable alarm identities, source-set mismatches, the current graph, its derived order, and prior outcomes. Structural validity and empirical utility remain determined by code and labeled data. Meaningful node names are not required because EvoCause can operate on arbitrary identifiers by interpreting source-set mismatches, graph structure, derived order, and prior outcomes. Human-readable alarm titles provide an additional signal when available.

The optimization history retains the last $K_{\mathcal{H}}$ rounds,
\begin{equation}
\mathbf{H}=\big\{(\mathcal{G}_{o-1},S_{o-1}),\ldots,(\mathcal{G}_{o-K_{\mathcal{H}}},S_{o-K_{\mathcal{H}}})\big\}.
\end{equation}

where $S_{o-1}=(L_{o-1},F_{o-1})$ is the lexicographic score of graph $\mathcal{G}_{o-1}$ on the full alignment set $\Dalign$. Here, $L_{o-1}$ is Case EM and $F_{o-1}$ is Node F1, which is used only to break ties. Specifically,
\begin{equation}
L_{o-1}=\frac{1}{N}\sum_{i=1}^{N}\indic\!\left\{\RCA(\mathbf{s}_i,\mathcal{G}_{o-1})=y_i\right\}.
\end{equation}
Thus, model selection first maximizes exact root-set recovery over the full alignment set and uses partial node-level recovery only when two candidate graphs have equal Case EM.

\emph{Step 4: Validate and apply edits.} The parser first checks that every operation is in $\{\emph{add},\emph{remove},\emph{reverse}\}$ and that both endpoints belong to $\mathbf{A}$. Valid edits are applied to obtain a tentative graph
\begin{equation} 
\mathcal{G}'=\ApplyG(\mathcal{G},\Delta\mathcal{G}). 
\end{equation}
The tentative graph is accepted if and only if $\mathcal{G}'$ remains a DAG. Otherwise, the complete proposed update is rejected and the current graph is retained. Any topological order needed by the next prompt is then recomputed deterministically from the accepted graph, so no graph order synchronization state or compatibility check is required.

These checks guarantee structural feasibility, not causal correctness. The refinement loop may visit several structurally valid candidate graphs. Following each round, EvoCause evaluates the current graph on the labeled alignment set and stores the graph with the highest Case EM, using Node F1 only as a tie-breaker.

\emph{Termination.} The four steps repeat across batches and epochs until either all $N_e$ epochs are processed or the round counter reaches the budget $N_r$. To improve efficiency, the procedure also terminates early if the graph remains unchanged for $c_{\mathrm{stop}}$ consecutive batches. The graph with the highest Case EM on the alignment set is returned as $\mathcal{G}^{*}$ and used for all downstream RCA. The full procedure is summarized in Appendix~\ref{app:optimization}.
 
\subsection{Inference Stage}
\label{sec:inference}

At test time, a new incident $\mathbf{s}_{\mathrm{new}}\in\mathcal{S}_{\mathrm{on}}$ is resolved using the refined graph $\mathcal{G}^{*}$ alone. The procedure identifies the alarm types present in $\mathbf{s}_{\mathrm{new}}$, restricts $\mathcal{G}^{*}$ to the subgraph they induce, and returns the complete set of nodes whose in-degree in this induced subgraph is zero. These source alarm types constitute the predicted root-cause set. Formally,
\begin{equation}
\hat{y}_{\mathrm{new}}=\RCA(\mathbf{s}_{\mathrm{new}},\mathcal{G}^{*}).
\end{equation}
This is the same procedure used to score candidate graphs during learning, so the criterion optimized offline is exactly the one applied online. No LLM call is made at inference time. Every prediction therefore traces back to specific edges in $\mathcal{G}^{*}$, and every such edge is attributable either to the initial discovery stage or to a recorded LLM graph edit during refinement, providing an audit trail that operators can inspect. Expert labels are used only during offline alignment to score candidate graphs and construct mismatch evidence. Their effect is retained in $\mathcal{G}^{*}$, so held-out incidents require neither labels nor LLM calls.

%% file: Sub_sections/Experiments.tex
\section{Experiments}
\label{sec:experiment}

We evaluate EvoCause on synthetic networks with known causal graphs and on \dataset, an expert-labeled production telecommunication dataset. Synthetic experiments assess RCA, graph reconstruction, and backbone sensitivity using non-semantic node identifiers, while TeleRCA evaluates RCA, robustness to causal-discovery initialization, and the contribution of human-readable alarm titles.

\subsection{Experimental Setup}

For both datasets, we use non-overlapping 80\% training and 20\% test splits. Each initial graph is learned only from the training split. Within that split, 10\% of the incidents form the labeled alignment set used for graph refinement and model selection. The held-out test split is evaluated once after the best alignment-set graph has been selected.

\paragraph{Evaluation metrics.} We evaluate root cause analysis on both datasets using Node F1 and Case EM. Let $\widehat{y}_i$ and $y_i$ denote the predicted and true root cause sets for test incident $i$, and let $N$ be the number of test incidents. Node F1 measures micro-averaged node-level overlap, while Case EM, where EM denotes exact match, measures incident-level accuracy:
\begin{equation} \resizebox{0.98\columnwidth}{!}{$\displaystyle \mathrm{Node\ F1}=\frac{2\sum_{i=1}^{N}|\widehat{y}_i\cap y_i|}{\sum_{i=1}^{N}\left(|\widehat{y}_i|+|y_i|\right)}, \qquad \mathrm{Case\ EM}=\frac{1}{N}\sum_{i=1}^{N}\mathbf{1}\!\left[\widehat{y}_i=y_i\right]$}. \end{equation}
Node F1 gives partial credit for recovering part of the true root set, whereas Case EM counts an incident as correct only when the complete predicted and true sets are identical. On synthetic data, the known ground-truth graph also allows evaluation of graph reconstruction. Let $\widehat{E}$ and $E$ denote the estimated and true directed edge sets:
\begin{equation} \mathrm{Graph\ F1}=\frac{2|\widehat{E}\cap E|}{|\widehat{E}|+|E|}, \qquad \mathrm{nSHD}=\frac{\mathrm{SHD}(\widehat{G},G)}{|E|}. \end{equation}
Graph F1 measures directed-edge recovery. Here, $\mathrm{SHD}(\widehat{G},G)=N_{\mathrm{add}}+N_{\mathrm{del}}+N_{\mathrm{rev}}$, where the three terms are the minimum numbers of unit-cost edge additions, deletions, and reversals required to transform $\widehat{G}$ into $G$. Higher Node F1, Case EM, and Graph F1 are better, while lower nSHD is better. Because \dataset\ has no ground-truth propagation graph, graph-reconstruction metrics are reported only on synthetic data.

\paragraph{Baselines and controlled variants.} Across both datasets, we compare unrefined graphs learned by PC, NOTEARS, and THPs with Chain-of-Event~\citep{yao2024coe}, CCCM~\citep{zhang2024cluster}, APGNN~\citep{jiang2023apgnn}, and RUN~\citep{lin2024run}, using the released default settings for the latter four methods. For PC and NOTEARS, we follow the alarm-transaction representation of~\citet{zhang2021influence}. Random uses a DAG with matched node and edge counts on synthetic data and a random alarm ordering on \dataset. EvoCause uses Qwen3-32B with alignment fraction $\rho=0.1$, batch size $N_b=20$, epoch count $N_e=10$, round budget $N_r=500$, history window $K_{\mathcal{H}}=5$, and retry budget $N_{\mathrm{retry}}=3$ per batch. Hyperparameter robustness and label-feedback efficiency are examined in Appendices~\ref{app:hyperparameter} and~\ref{app:feedback_efficiency}, respectively. On synthetic data, two Qwen3-4B variants assess backbone sensitivity only~\citep{yang2025qwen3}. On TeleRCA, EvoCause-PC (Anonymous IDs) replaces alarm titles with fixed anonymous identifiers while keeping all other settings unchanged to test the contribution of human-readable titles.

\subsection{Synthetic Data}

\paragraph{Dataset.} Synthetic data allow evaluation against fully known graphs. We generate ten independent Erd\H{o}s--R\'{e}nyi  DAGs, each with 50 alarm-type nodes and 70 directed edges before transitive reduction. The reduction preserves all nodes but removes shortcut edges. We retain 2,000 valid incidents per graph, with one to five root alarms per incident. Each active edge generates one offspring event after an exponential delay whose edge-specific mean is sampled once from $[30,300]$ seconds and fixed across incidents. We report the mean and standard deviation across the ten graphs. Because alarm types use numeric identifiers, these experiments evaluate structural refinement without alarm-name semantics.

\begin{table*}[t]
\centering
{\footnotesize
\setlength{\tabcolsep}{3pt}
\renewcommand{\arraystretch}{0.92}
\begin{tabular}{@{}lcccc@{}}
\toprule
\textbf{Method} & Node F1 (\%) $\uparrow$ & Case EM (\%) $\uparrow$ & Graph F1 (\%) $\uparrow$ & nSHD $\downarrow$ \\
\midrule
Random & $47.23\pm8.42$ & $6.05\pm3.18$ & $5.71\pm4.89$ & $1.8714\pm0.1421$ \\
PC & $52.03\pm6.17$ & $17.80\pm4.48$ & $64.80\pm21.58$ & $0.7859\pm0.7483$ \\
NOTEARS & $30.03\pm2.95$ & $7.43\pm2.38$ & $34.38\pm11.29$ & $0.8593\pm0.1554$ \\
THPs & $37.05\pm6.54$ & $10.60\pm3.48$ & $47.93\pm4.12$ & $0.6882\pm0.0317$ \\
\midrule
Chain-of-Event~\citep{yao2024coe} & $53.45\pm8.11$ & $22.98\pm6.29$ & $53.71\pm4.92$ & $1.4795\pm0.2141$ \\
CCCM~\citep{zhang2024cluster} & $58.10\pm7.11$ & $25.43\pm5.51$ & $46.23\pm6.98$ & $1.6179\pm0.3218$ \\
APGNN~\citep{jiang2023apgnn} & $61.13\pm7.60$ & $24.17\pm6.25$ & $51.99\pm3.94$ & $1.5741\pm0.1745$ \\
RUN~\citep{lin2024run} & $60.80\pm4.69$ & $23.30\pm4.30$ & $54.67\pm8.55$ & $1.2931\pm0.2995$ \\
\midrule
EvoCause-PC (4B-Inst.) & $57.78\pm6.03$ & $21.92\pm5.41$ & $66.61\pm14.86$ & $0.6887\pm0.5529$ \\
EvoCause-PC (4B-Think.) & $62.80\pm6.72$ & $26.62\pm6.98$ & $68.07\pm10.63$ & $0.5673\pm0.2100$ \\
EvoCause-PC (32B) & $\bm{63.62\pm6.61}$ & $\bm{27.20\pm6.51}$ & $\bm{69.39\pm8.76}$ & $\bm{0.5480\pm0.1462}$ \\
EvoCause-NOTEARS (32B) & $55.09\pm6.40$ & $20.95\pm4.68$ & $37.94\pm8.57$ & $0.7837\pm0.0645$ \\
EvoCause-THPs (32B) & $55.74\pm7.24$ & $20.75\pm4.41$ & $50.32\pm3.20$ & $0.6645\pm0.0283$ \\
\bottomrule
\end{tabular}
}
\caption{Synthetic-data results on ten transitively reduced Erd\H{o}s--R\'{e}nyi DAGs. Values are mean $\pm$ standard deviation across graph instances. 32B, 4B-Inst., and 4B-Think. denote Qwen3-32B, Qwen3-4B-Instruct-2507, and Qwen3-4B-Thinking-2507. Full component and backbone analyses are reported in Appendices~\ref{app:component_ablation} and~\ref{app:backbone_sensitivity}.}
\label{tab:synthetic_data_v3}
\end{table*}

\paragraph{Analysis.} Table~\ref{tab:synthetic_data_v3} shows that label-feedback refinement improves RCA for all three initial discovery methods, so the benefit is not tied to the inductive bias of PC. With PC initialization, EvoCause raises Node F1 and Case EM by $11.59$ and $9.40$ percentage points, increases Graph F1 by $4.59$ percentage points, and reduces nSHD by $0.2379$. RCA improves more strongly than complete edge recovery for NOTEARS and THPs, which is consistent with expert labels directly constraining incident-level source sets but only indirectly constraining the complete edge set. EvoCause-PC gives the strongest joint RCA and graph-reconstruction result, showing that refinement preserves the value of a strong initial graph. Because synthetic nodes use arbitrary numeric identifiers, these improvements also show that EvoCause can exploit source-set mismatches and graph context without relying on lexical information. The backbone comparisons evaluate sensitivity to the choice of LLM and demonstrate that the framework can operate with different LLM backbones.

\subsection{Real-World Telecommunication Data}
\label{sec:real}

\paragraph{Dataset.} \dataset\ contains $10{,}922$ incidents comprising $485{,}681$ alarm events across $119{,}039$ timestamps, $5{,}621$ network resources, and $194$ alarm types, collected from a production telecommunication network in Jakarta, Indonesia, between March and July 2025. Potentially sensitive information was anonymized to meet privacy and publication requirements. The incident-file structure and field definitions are provided in Appendix~\ref{app:telerca_structure}.

\begin{table}[h]
\centering
{\footnotesize
\setlength{\tabcolsep}{3pt}
\renewcommand{\arraystretch}{0.92}
\begin{tabular}{@{}lcc@{}}
\toprule
\textbf{Method} & Node F1 (\%) $\uparrow$ & Case EM (\%) $\uparrow$ \\
\midrule
Random & $71.69\pm2.07$ & $62.15\pm2.79$ \\
PC & $65.18$ & $53.98$ \\
NOTEARS & $70.69$ & $61.03$ \\
THPs & $76.59$ & $68.86$ \\
\midrule
CCCM & $75.95$ & $67.35$ \\
Chain-of-Event & $65.82$ & $54.81$ \\
APGNN & $84.14\pm0.91$ & $78.30\pm1.23$ \\
RUN & $57.65\pm0.03$ & $45.03\pm0.05$ \\
\midrule
EvoCause-PC & $\bm{92.58\pm0.04}$ & $\bm{89.61\pm0.05}$ \\
EvoCause-PC (Anonymous IDs) & $86.46\pm0.06$ & $81.57\pm0.08$ \\
EvoCause-NOTEARS & $92.47\pm0.07$ & $89.44\pm0.10$ \\
EvoCause-THPs & $92.48\pm0.04$ & $89.46\pm0.05$ \\
\bottomrule
\end{tabular}
}
\caption{RCA performance on \dataset. Values are mean $\pm$ standard deviation over five runs where available. EvoCause-PC (Anonymous IDs) replaces the original alarm titles supplied to the LLM with fixed anonymous identifiers. All EvoCause variants use Qwen3-32B, whereas deterministic discovery baselines are single runs.}
\label{tab:real_data}
\end{table}

\paragraph{Analysis.} Table~\ref{tab:real_data} shows that EvoCause reaches nearly identical RCA performance from PC, NOTEARS, and THPs despite substantial differences among their unrefined results. With PC initialization, EvoCause improves Node F1 from $65.18\%$ to $92.58\%$ and Case EM from $53.98\%$ to $89.61\%$, gains of $27.40$ and $35.63$ percentage points. This convergence indicates that label-feedback alignment reduces dependence on the initial discovery bias. Replacing the original alarm titles with anonymous identifiers lowers Node F1 to $86.46\%$ and Case EM to $81.57\%$, reductions of $6.12$ and $8.04$ percentage points. Because all other settings remain unchanged, this result indicates that human-readable alarm-name information provides additional guidance for graph-edit proposals. The anonymized variant still exceeds the strongest non-EvoCause baseline by $2.32$ percentage points in Node F1 and $3.27$ percentage points in Case EM, showing that feedback-guided structural refinement remains effective without lexical information. Since \dataset\ has no ground-truth propagation graph, these results demonstrate diagnostic utility and robustness rather than exact structural recovery.

%% file: Sub_sections/Conclusion.tex
\section{Conclusion}

We release TeleRCA, an expert-annotated production benchmark, and propose EvoCause, a discovery-agnostic framework that uses historical expert diagnoses to refine alarm graphs. Because incident labels specify desired source nodes but not the required edge changes, an LLM proposes semantically plausible graph updates. Deterministic procedures validate node identities and acyclicity and retain the best graph on a labeled alignment set. Experiments on synthetic data and TeleRCA show that EvoCause improves root cause analysis over unrefined discovery graphs and other RCA baselines, while synthetic results also show improved graph reconstruction. The TeleRCA name-anonymization result further indicates that human-readable alarm-title information provides additional guidance for refinement. At inference, the refined graph alone produces predictions without expert labels or LLM calls. Since root-cause labels generally identify a root-cause-consistent graph set rather than a unique data-generating DAG, future work will strengthen identifiability by incorporating intervention records, temporal evidence, and expert edge-level constraints. We also plan to model resource-specific and time-varying propagation, account for latent common causes, and improve robustness to noisy expert labels through active selection of informative incidents.

%% file: Sub_sections/Appendix.tex
\section{Appendix}
\label{sec:appendix}

\subsection{EvoCause Optimization Procedure}
\label{app:optimization}

Algorithm~\ref{alg:evocause} compares candidate graphs using the lexicographic score $S=(\mathrm{CaseEM},\mathrm{NodeF1})$, prioritizing exact incident resolution and using partial node recovery only to break ties. At epoch $e$, $\Dalign^{(e)}$ is a random permutation of $\Dalign$ used to form the batches. The graph is the only persistent structural state, while its topological order is recomputed for each prompt. The LLM returns only graph updates, invalid or cyclic candidates are rejected, and the graph with the best alignment-set score is retained.

\begin{algorithm}[h]
\caption{EvoCause: LLM-Guided Evolution of Causal Graphs for RCA}
\label{alg:evocause}
\begin{algorithmic}[1]
\REQUIRE Historical sequences $\Dhist$, labeled alignment cases $\Dalign=\{(s_i,y_i)\}_{i=1}^{N}$, discovery algorithm $\CD$, LLM $\mathcal{M}$, prompt $p_u$, batch size $N_b$, epochs $N_e$, round budget $N_r$, patience $c_{\mathrm{stop}}$, seed $\xi$
\ENSURE Refined graph $\mathcal{G}^{*}$
\STATE $\mathcal{G}\gets\CD(\Dhist)$
\STATE $S^{*}\gets\mathrm{Score}(\Dalign,\mathcal{G})$\quad $\mathcal{G}^{*}\gets\mathcal{G}$
\STATE $r\gets0$\quad $c\gets0$\quad $\mathcal{H}\gets\emptyset$
\FOR{$e=1$ \TO $N_e$}
    \STATE $\Dalign^{(e)}\gets\mathrm{Shuffle}(\Dalign;\xi+e)$
    \STATE $\{\mathcal{B}_k\}_{k=1}^{K}\gets\mathrm{Partition}(\Dalign^{(e)},N_b)$
    \FOR{$k=1$ \TO $K$}
        \STATE $\mathcal{G}_{\mathrm{prev}}\gets\mathcal{G}$
        \STATE $\mathcal{E}\gets\{(s_i,\RCA(s_i,\mathcal{G}),y_i):(s_i,y_i)\in\mathcal{B}_k\}$
        \STATE $O(\mathcal{G})\gets\TopoOrder(\mathcal{G})$
        \STATE $\Delta\mathcal{G}\gets\mathcal{M}(\mathcal{G},O(\mathcal{G}),\mathcal{E},\mathcal{H},p_u)$
        \STATE $\mathcal{G}'\gets\ApplyG(\mathcal{G},\Delta\mathcal{G})$
        \IF{$\mathrm{ValidUpdates}(\Delta\mathcal{G},\mathbf{A})$ \AND $\mathrm{IsDAG}(\mathcal{G}')$}
            \STATE $\mathcal{G}\gets\mathcal{G}'$
        \ELSE
            \STATE Reject the update and keep $\mathcal{G}$ unchanged
        \ENDIF
        \STATE $S\gets\mathrm{Score}(\Dalign,\mathcal{G})$
        \IF{$S>_{\mathrm{lex}}S^{*}$}
            \STATE $S^{*}\gets S$\quad $\mathcal{G}^{*}\gets\mathcal{G}$
        \ENDIF
        \STATE $c\gets c+1$ if $\mathcal{G}=\mathcal{G}_{\mathrm{prev}}$, otherwise $c\gets0$
        \STATE Append $(\mathcal{G},S)$ to $\mathcal{H}$, then set $r\gets r+1$
        \IF{$r\geq N_r$ \OR $c\geq c_{\mathrm{stop}}$}
            \RETURN $\mathcal{G}^{*}$
        \ENDIF
    \ENDFOR
\ENDFOR
\RETURN $\mathcal{G}^{*}$
\end{algorithmic}
\end{algorithm}
\FloatBarrier

\subsection{TeleRCA Data Structure}
\label{app:telerca_structure}

Each \dataset{} incident is stored as one JSON file with two main fields: \texttt{nodes} contains the resources and alarms observed during the incident, and \texttt{causeNode} contains the expert annotation. Each alarm record includes an identifier, a human-readable title, and a UTC timestamp. The resource identifier is retained in the released record but is not part of the event triplet $(a_j,t_j,l_j)$ used in the method formulation.

Each key in \texttt{causeNode} identifies a root-cause resource and maps to one or more alarm identifiers. These identifiers are matched to alarm records under the corresponding entry in \texttt{nodes}. The matched alarm titles define the expert root set, and the remaining alarms are downstream symptoms. The abbreviated example below contains one labeled \texttt{Physical Port Down} root alarm.

\begin{lstlisting}[
  basicstyle=\ttfamily\scriptsize,
  breaklines=true,
  columns=fullflexible,
  frame=single
]
{"nodes":[
  {"device":"JKT-GATG-OPT-H910D",
   "alarms":[{"id":"a1","title":"BFD session down","t":"03:22:15"}]},
  {"device":"JKT-RJG-AN1-H8X08",
   "alarms":[{"id":"a2","title":"Physical Port Down","t":"03:22:04"},
             {"id":"a3","title":"Link Down","t":"03:22:04"}]}],
 "causeNode":{"JKT-RJG-AN1-H8X08":["a2"]}}
\end{lstlisting}

\begin{table*}[!t]
\centering
{\small
\setlength{\tabcolsep}{4pt}
\begin{tabular}{@{}lcccc@{}}
\toprule
\textbf{Variant or backbone} & Node F1 (\%) $\uparrow$ & Case EM (\%) $\uparrow$ & Graph F1 (\%) $\uparrow$ & nSHD $\downarrow$ \\
\midrule
\multicolumn{5}{@{}l}{\textit{Component ablations}} \\
w/o Topological Order Context & $60.72\pm6.84$ & $22.19\pm5.56$ & $68.44\pm16.02$ & $0.6231\pm0.4029$ \\
w/o Optimization History & $62.81\pm6.46$ & $26.33\pm6.87$ & $68.69\pm12.20$ & $0.5746\pm0.2558$ \\
EvoCause & $\bm{63.62\pm6.61}$ & $\bm{27.20\pm6.51}$ & $\bm{69.39\pm8.76}$ & $\bm{0.5480\pm0.1462}$ \\
\midrule
\multicolumn{5}{@{}l}{\textit{LLM backbone sensitivity}} \\
Qwen3-4B-Instruct-2507 & $57.78\pm6.03$ & $21.92\pm5.41$ & $66.61\pm14.86$ & $0.6887\pm0.5529$ \\
DeepSeek-R1-Distill-Llama-70B & $61.18\pm6.63$ & $25.53\pm6.46$ & $67.70\pm12.26$ & $0.5918\pm0.3591$ \\
Qwen3-4B-Thinking-2507 & $62.80\pm6.72$ & $26.62\pm6.98$ & $68.07\pm10.63$ & $0.5673\pm0.2100$ \\
Qwen3-32B & $\bm{63.62\pm6.61}$ & $\bm{27.20\pm6.51}$ & $\bm{69.39\pm8.76}$ & $\bm{0.5480\pm0.1462}$ \\
\bottomrule
\end{tabular}
}
\caption{Component ablations and LLM-backbone sensitivity on the ten synthetic graph instances with PC initialization. The component study uses Qwen3-32B except for the stated variant. Values are mean $\pm$ standard deviation.}
\label{tab:ablation}
\end{table*}

\subsection{Synthetic Component Ablations}
\label{app:component_ablation}
Table~\ref{tab:ablation} reports the component-ablation results. We initialize EvoCause with PC, use Qwen3-32B as the default proposal model, and change one prompt component at a time. \textsc{w/o History} removes prior graph--score pairs, while \textsc{w/o Order} removes the topological linearization derived from the current graph. These variants evaluate the contribution of structural prompt context on synthetic graphs whose nodes have arbitrary numeric identifiers.

Removing the derived order causes the larger degradation, lowering Node F1, Case EM, and Graph F1 by $2.90$, $5.01$, and $0.95$ percentage points, respectively, while increasing nSHD by $0.0751$. Removing optimization history causes smaller reductions of $0.81$, $0.87$, and $0.70$ percentage points in Node F1, Case EM, and Graph F1, respectively, while increasing nSHD by $0.0266$. These results show that both structural context components improve refinement, with the derived topological order having the larger overall contribution.

\subsection{LLM Backbone Sensitivity}
\label{app:backbone_sensitivity}
We fix the initial PC graph, prompt, alignment cases, graph-edit budget, and deterministic validator, and vary only the LLM backbone. We compare Qwen3-4B-Instruct-2507, Qwen3-4B-Thinking-2507, DeepSeek-R1-Distill-Llama-70B, and the default Qwen3-32B~\citep{yang2025qwen3,deepseek2025r1}. Because these models may differ in several respects beyond parameter count, this experiment is treated as a backbone sensitivity analysis rather than an isolation of any specific model capability.

Among the backbone results in Table~\ref{tab:ablation}, Qwen3-32B obtains the best mean result on all four metrics, while performance varies across the remaining backbones. Because the models differ in multiple respects, this comparison does not isolate the cause of the observed differences. We therefore use it only to assess the sensitivity of EvoCause to the selected LLM backbone.

\subsection{Hyperparameter Robustness}
\label{app:hyperparameter}
\begin{figure*}[!t]
\centering
\includegraphics[width=0.94\textwidth]{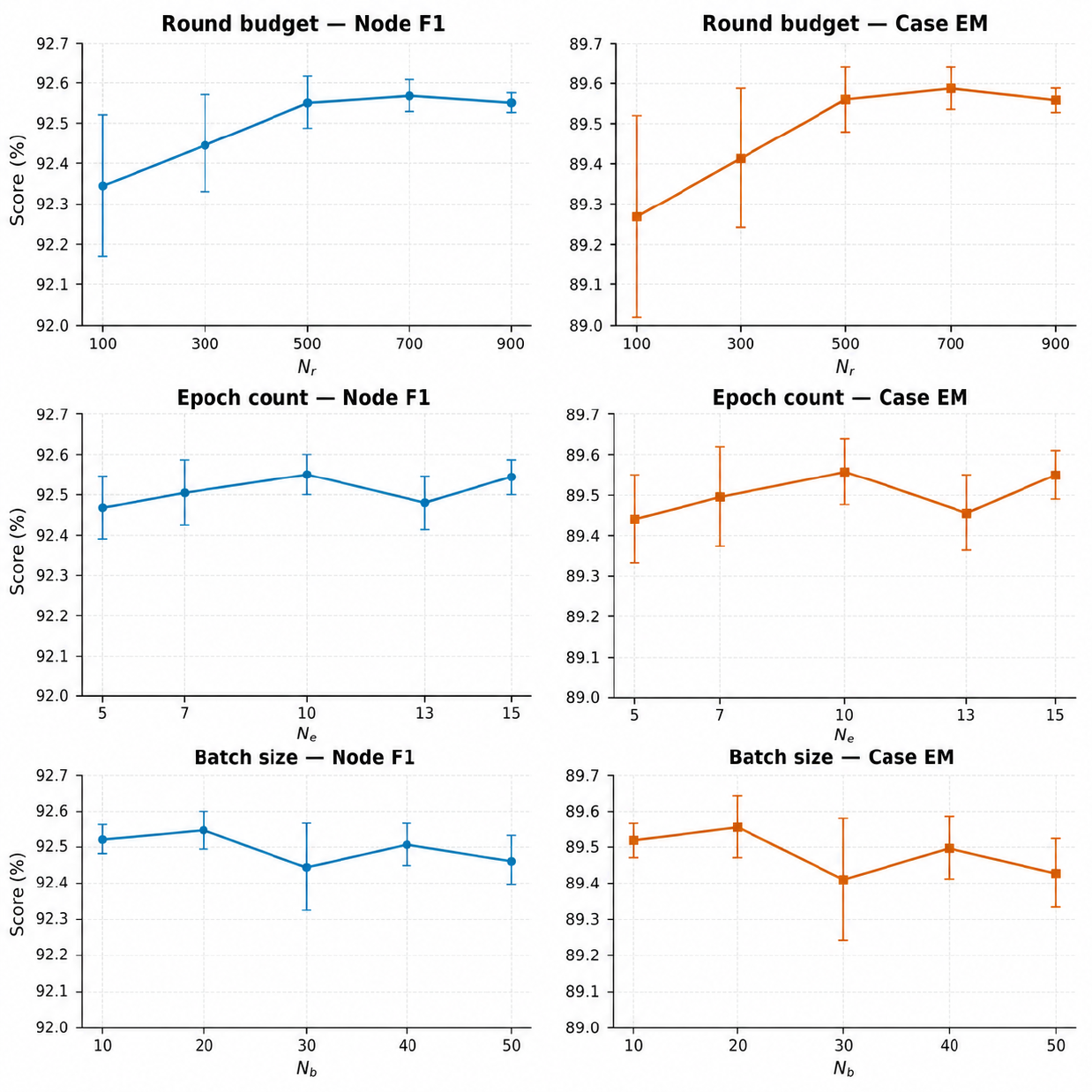}
\caption{Sensitivity to the refinement-round budget, epoch count, and batch size. Each point reports mean and standard deviation over five runs. The left and right columns show Node F1 and Case EM, respectively.}
\label{fig:hyper_parameter_varying}
\end{figure*}

Figure~\ref{fig:hyper_parameter_varying} summarizes the sensitivity of EvoCause to the round budget, epoch count, and batch size. On \dataset, incidents are partitioned 4:1 into non-overlapping train and test sets. The default values are $N_r=500$ refinement rounds, $N_e=10$ epochs, and $N_b=20$ cases per batch. We vary one parameter at a time, run each setting five times, and report Node F1 and Case EM.
\paragraph{Round budget.} For this robustness study, we use $20\%$ of the training split as the labeled alignment set, i.e., $\rho_{\mathrm{rob}}=0.2$, and vary the round budget over $N_r\in\{100,300,500,700,900\}$. Performance improves up to approximately $N_r=700$ and then stabilizes, while variability generally decreases as the budget increases.
\paragraph{Epoch count.} We vary $N_e\in\{5,7,10,13,15\}$. Both metrics remain within a narrow range, with a mild peak near the default $N_e=10$, indicating diminishing returns after a moderate number of shuffled passes through the alignment cases.
\paragraph{Batch size.} We vary $N_b\in\{10,20,30,40,50\}$. The default $N_b=20$ gives the best mean performance. Larger batches slightly degrade both metrics and increase variance, consistent with the greater number of cross-incident constraints that must be reconciled in one proposal.

\subsection{Label-Feedback Efficiency}
\label{app:feedback_efficiency}
\begin{center}
\begin{minipage}{0.96\columnwidth}
\centering
\includegraphics[width=\linewidth]{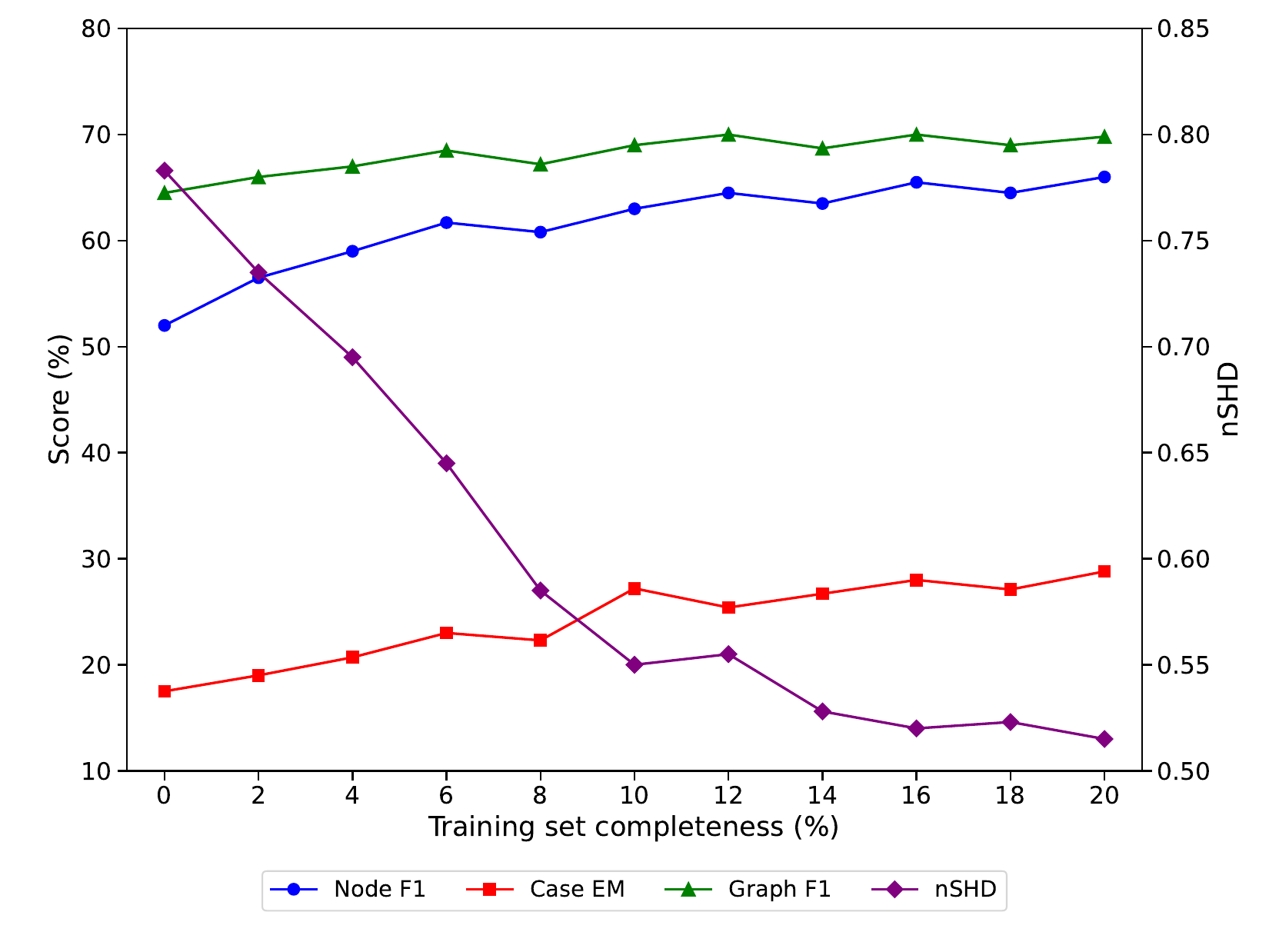}
\captionof{figure}{Synthetic-data performance under PC initialization as the labeled alignment fraction increases from 0\% to 20\%. The x-axis label ``training set completeness'' denotes the percentage of the training split used for label-feedback alignment.}
\label{fig:alignment_fraction}
\end{minipage}
\end{center}
 
Figure~\ref{fig:alignment_fraction} varies the percentage of the training split used as the labeled alignment set from 0\% to 20\% in steps of two percentage points under PC initialization. The 0\% point is the unrefined PC graph. All four metrics improve sharply once a small labeled subset is introduced. Node F1, Case EM, and Graph F1 then fluctuate within a comparatively narrow range, whereas nSHD continues to decline as more labels constrain the admissible graph set. The default 10\% setting lies in this stable region and balances annotation cost with refinement quality.